%% file: main.tex
\documentclass[letterpaper]{article} 
\usepackage{aaai24}  
\usepackage{times}  
\usepackage{helvet}  
\usepackage{courier}  
\usepackage[hyphens]{url}  
\usepackage{graphicx} 
\usepackage{natbib}  
\usepackage{caption} 
\usepackage{tabularx}

\usepackage{algorithm}
\usepackage{algorithmic}

\usepackage{newfloat}
\usepackage{listings}
\DeclareCaptionStyle{ruled}{labelfont=normalfont,labelsep=colon,strut=off} 
\floatstyle{ruled}
\newfloat{listing}{tb}{lst}{}
\floatname{listing}{Listing}
\usepackage{color} 

\title{Utilising Large Language Models for Generating Effective Counter Arguments to Anti-Vaccine Tweets}

\author{
   Utsav Dhanuka\textsuperscript{\rm $\S$}\equalcontrib,
   Soham Poddar\equalcontrib,
   Saptarshi Ghosh
}
\affiliations{
   Indian Institute of Technology Kharagpur, West Bengal 721302, India\\
   
   \textsuperscript{\rm $\S$} utsavdhanuka03@kgpian.iitkgp.ac.in   
}

\usepackage{multirow}
\usepackage{amsmath}
\usepackage{amssymb}
\usepackage{dsfont}
\usepackage{subcaption}
\usepackage{tikz}
\usepackage{soul}
\usepackage{adjustbox}
\usepackage{booktabs}
\usepackage{makecell}
\usepackage{subcaption}   
\usepackage{caption}      
\newcommand{\counterdata}[0]{CNTR-VAX}

\usepackage{xcolor}

\begin{document}









\maketitle

\begin{abstract}
    \input{sections/abstract}
\end{abstract}


\input{sections/introduction}
\input{sections/relatedwork}
\input{sections/generatingarguments}

\input{sections/evaluation}

\input{sections/twostepcot}

\input{sections/knowledgedistill}

\input{sections/conclusion}



\nocite{islam2024latent}
\nocite{saha2024misinfo}
\nocite{alshomary2021argument}
\nocite{furman2024mining}
\nocite{germani2022how}


\bibliography{ref}

\appendix
\input{sections/appendix}

\end{document}

%% file: sections/abstract.tex
In an era where public health is increasingly influenced by information shared on social media, combatting vaccine skepticism and misinformation has become a critical societal goal. Misleading narratives around vaccination have spread widely, creating barriers to achieving high immunization rates and undermining trust in health recommendations. While efforts to detect misinformation have made significant progress, the generation of real-time counter-arguments tailored to debunk such claims remains an underexplored area. In this work, we explore the capabilities of LLMs to generate sound counter-argument rebuttals to vaccine misinformation. Building on prior research in misinformation debunking, we experiment with various prompting strategies and fine-tuning approaches to optimize counter-argument generation. Additionally, we train classifiers to categorize anti-vaccine tweets into multi-labeled categories—such as concerns about vaccine efficacy, side effects, and political influences—allowing for more context-aware rebuttals. Our evaluation, conducted through human judgment, LLM-based assessments, and automatic metrics, reveals strong alignment across these methods. Our findings demonstrate that integrating label descriptions and structured fine-tuning enhances counter-argument effectiveness, offering a promising approach for mitigating vaccine misinformation at scale.

%% file: sections/introduction.tex
\section{Introduction}
The rapid proliferation of social media platforms such as Twitter and Reddit has
significantly transformed the way information is disseminated and consumed. However,
this transformation has also facilitated the spread of misinformation, with
anti-vaccine rhetoric becoming a prominent example. Such misinformation can erode
public trust in vaccines, impede public health efforts, and contribute to the resurgence
of preventable diseases. Addressing these challenges requires innovative solutions
that go beyond manual moderation and fact-checking due to the sheer scale of
content generated daily.
The COVID-19 pandemic, which began in 2020, has underscored the critical importance
of mass vaccination as the most effective means to achieve herd immunity and
mitigate the virus’s impact. Despite the availability of safe and effective vaccines,
misinformation campaigns have fueled skepticism and resistance, endangering public
health initiatives worldwide. To counter this, there is an urgent need for automated
systems capable of responding to anti-vaccine claims with persuasive and factual
counter-arguments in real time.

While efforts to detect misinformation have made significant progress, the generation
of real-time counter-arguments tailored to debunk such claims remains an
underexplored area. Our works seeks to bridge that gap by focusing on the automatic
generation of counter-arguments for anti-vaccine tweets. By providing timely,
informative, and convincing rebuttals, we can reduce the spread of misinformation
and promote trust in vaccination efforts. Prior studies have explored the
theoretical underpinnings of countering anti-vaccination attitudes, highlighting both
cognitive mechanisms and dual-process theories to address resistance to vaccination
messages \cite{okuhara2020dual,horne2015countering}. However, these works largely focus on the psychology of persuasion
and broad message framing rather than on the generation of targeted counterarguments
addressing specific user concerns. To the best of our knowledge, our work
is the first to systematically tackle the problem of generating fine-grained, concernspecific
counter-arguments for anti-vaccine content using large language models.

In this work we focus on developing a comprehensive approach to generate
counter-arguments for anti-vaccine tweets by leveraging recent advancements in
artificial intelligence and natural language processing. Key contributions of our work include:
\begin{itemize}
    \item \textbf{First}, we establish an effective strategy for addressing anti-vaccine rhetoric through prompt engineering techniques applied to GPT-4o-Mini. To ensure the generated counter-arguments are both relevant and impactful, we employ a rigorous evaluation framework that incorporates human judgment, LLM-based assessments, and automatic metrics to assess dimensions such as factual accuracy, clarity, coverage, and persuasiveness.

    \item  \textbf{Second}, we develop classifiers capable of generating label descriptions for anti-vaccine tweets, which are then used to prompt smaller language models (SLMs) for producing effective and context-aware rebuttals. We further enhance this process by designing a Chain-of-Thought (CoT) reasoning approach, where the model is guided to first infer relevant concerns or themes before generating the counter-argument, leading to more coherent and targeted responses.


    \item \textbf{Finally}, we fine-tune small language models (SLMs) on high-quality counter-arguments generated by GPT-4o to enable them to operate independently. These models are evaluated using standard automatic metrics such as ROUGE and BERTScore. Our approach highlights the practicality of using SLMs, which offer a favorable trade-off between performance and computational efficiency, making them suitable for deployment in real-world, resource-constrained settings.
    
\end{itemize}

%% file: sections/relatedwork.tex
\section{Related Work}
In this section, we review prior research on counter-argument generation, highlighting its significance in online discourse and exploring how effective counter-arguments should be constructed. We also discuss related efforts in countering misinformation, hate speech, and other adversarial content using language models and prompting strategies.

\vspace{2mm}
\noindent\textbf{Approaches to Counter-Narratives:}
Recent work has shown that techniques developed for political argumentation and counterspeech can inform strategies to counter vaccine misinformation effectively. 
\citet{palmer2023llmPolitics} demonstrate that Large Language Models (LLMs) are capable of producing persuasive arguments on politically polarizing topics that are often as convincing as human-authored ones. However, they also find that human evaluators display bias against LLM-generated arguments when informed of their origin, highlighting the challenges of trust in AI-generated persuasion. Extending LLM capabilities to toxic online discourse, \citet{saha2022countergedi} propose COUNTERGEDI, a controllable generation framework that produces counterspeech with specific attributes such as politeness, emotional tone, and detoxification. Their model significantly improves the quality and acceptability of counterspeech, offering a promising avenue for generating nuanced responses to online hate, which is often closely intertwined with misinformation. Similarly, \citet{saha2024misinfo} propose a hybrid framework that adapts counter-hate strategies to misinformation domains, combining annotated response strategy classifiers with an argument-graph-guided LLM generator. Their approach demonstrates improvements in factual accuracy, naturalness, and persuasiveness.

\citet{alshomary2021argument} proposed argument undermining, a novel approach that identifies weak premises in arguments and generates responses that directly challenge those specific points, demonstrating improved quality in generated counter-arguments.  \citet{saha2024zeroshot} investigated zero-shot counterspeech generation using models like GPT-2, DialoGPT, ChatGPT, and Flan-T5, highlighting the efficacy of strategic prompting and model scaling while raising concerns around toxicity. A complementary study by \citet{islam2024latent} introduced an LLMs-in-the-loop framework to uncover latent arguments from social media discourse, which supports dynamic argument discovery in evolving domains such as COVID-19 vaccine hesitancy and climate change. 
These works collectively provide useful methods that inform our LLM-based counter-argument generation pipeline for addressing anti-vaccine discourse, ranging from premise analysis and dataset curation to prompting strategies and model-assisted argument extraction.

\noindent \textbf{Studies on Vaccine-related Arguments:}
Recent studies provide crucial insights into the psychological and communicative strategies needed to construct effective counter-arguments against vaccine misinformation. \citet{okuhara2020dual}  emphasize that anti-vaccine rhetoric often appeals to fast, emotional reasoning (System 1), while pro-vaccine messages typically rely on slower, analytical thinking (System 2). They argue that persuasive messages should engage both systems—combining emotional appeal with factual evidence. Complementing this, \citet{germani2022how} stress the importance of filling ``information voids'' where misinformation thrives, advocating for proactive public health messaging and increased digital literacy to build long-term resilience. \citet{horne2015countering} offer another important perspective, focusing on how the framing of counter-arguments influences their effectiveness. Their work reveals that direct myth-debunking strategies—such as presenting scientific facts that contradict anti-vaccine claims—often fail to change minds and may even backfire. Instead, they demonstrate that highlighting the real-world consequences of vaccine refusal, especially through moral and empathetic narratives (e.g., risks to children and communities), can significantly shift attitudes. 
Inspired by these results, our work uses large language models (LLMs) to produce factually sound, emotionally compelling, and context-sensitive counterarguments. In order to align with these evidence-based strategies—which include emotional framing, relevance, and clarity to enhance the persuasive quality of automated counter-narratives—we plan to condition the models on descriptive labels that capture misinformation themes.

\vspace{1mm}
\noindent \textbf{Studies on Vaccine misinformation:} 
\citet{poddar2022caves} introduced the CAVES dataset, a multi-label annotated collection of anti-vaccine tweets, offering rich explanations and summaries to facilitate explainable classification and summarization—laying the groundwork for targeted counter-argument tasks. Another work by \cite{Poddar_Mondal_Misra_Ganguly_Ghosh_2022} explored how vaccine-related stances on Twitter evolved over time, particularly across the pre-COVID and COVID periods. They developed a high-precision classifier (97\%) to categorize users based on their vaccine stance, and investigated shifts in discourse and topical trends within and across different user groups. Notably, their work also identified users who changed stance during the pandemic and analyzed potential drivers behind such shifts—offering valuable sociolinguistic context to vaccine discourse.
\cite{loomba2021measuring} also showed how exposure to false information about the COVID-19 vaccine can drastically lower vaccination intent in the United Kingdom and the United States, highlighting the practical repercussions of such content and the pressing need for efficient counter-misinformation tactics like focused counterargument creation.
However, to the best of our knowledge, no prior work has tried to automatically generate counter arguments to vaccine misinformation. 

%% file: sections/generatingarguments.tex
\section{Generating Counter Arguments} \label{sec:Generating Counter Arguments}

\subsection{Motivation for using LLM Generated Arguments}
To assess the viability of using large language models (LLMs) for countering vaccine misinformation at scale, we first compared their responses with those written by humans. We performed a pilot study to determine whether LLM-generated counterarguments could equal or even outperform human-crafted responses in several crucial areas of persuasive argumentation. Table \ref{Table: human_annotations} shows an example of counter-argument generated by humans and LLMs.

We selected a set of 40 anti-vaccine tweets from the CAVES dataset, Since expert annotators are very expensive to obtain we select two humans with average expertise in this field to write counter-arguments to the tweets. In order to produce corresponding counterarguments, the same tweet set was also fed into a number of LLMs, namely GPT-4o-mini, Phi-3, Tiny-LLaMA, and LLaMA-3.2. Using the human responses as the ground truth, we evaluated all outputs using BERTScore to capture the semantic similarity between the human responses and LLM responses. Since there were two humans, we took the average scores, which are displayed in Table \ref{tab:human_llm_comparison}

 We performed a qualitative analysis along four crucial dimensions—coverage, clarity, persuasiveness, and factualness—in addition to quantitative metrics. Our analysis suggested that LLM-generated responses demonstrated promising quality and, in some cases, approached or surpassed human-written ones. Their responses were more elaborate and complete as compared to the humans and they show good semantic similarity with human responses. The strongest performance was observed with GPT-4o, which produced thorough, factually sound, and persuasive counter-arguments. Interestingly, even smaller-scale models such as Phi-3, Tiny-LLaMA, and LLaMA-3.2 frequently outperformed human annotators in terms of factual accuracy and structural coherence. Human responses, while occasionally ethically grounded and factually accurate tended to show less consistency and depth overall. These findings suggest that contemporary LLMs, when guided by effective prompting strategies, hold strong potential for generating persuasive and educational counter-speech against vaccine misinformation.\\
Recent work by \cite{guo2024evaluatinglargelanguagemodels} has highlighted the potential of LLMs not only as direct classifiers but also as tools for augmenting human-annotated datasets, with their outputs often serving as strong reference standards. Building on this idea, we employ GPT-4o, a state-of-the-art LLM, to generate a sizable dataset of counter-arguments. We then conduct detailed linguistic, rhetorical, and factual analyses to assess the reliability and effectiveness of these arguments in addressing diverse forms of vaccine misinformation.

\input{tables/humanresponses}

In this section we discuss the creation and validation of the synthetic dataset \textbf{\counterdata{}} for countering anti-vaccine arguments.

\subsection{Dataset Preparation}

\input{tables/labeldesc}

For selecting tweets/posts to counter, we leverage the CAVES dataset~\cite{poddar2022caves} that contains about 10k anti-vaccine tweets, labelled with the specific vaccine-concerns in a multi-label fashion. 
These 12 concern classes are described in Table~\ref{tab:labels_map}.
We use OpenAI's GPT-4o-mini~\footnote{https://platform.openai.com/} model to generate counter arguments to tweets from the CAVES dataset.
We investigated two prompting strategies for generating counter argument to anti-vaccine tweets as described below.

\vspace{2mm}
\noindent \textbf{(a) Without Label Information:}
One method was to use standard prompts like ``\texttt{Generate a strong counter-argument for the tweet: [text]}'' to generate counterarguments. 
The produced counterarguments were of good quality (as discussed later in this section), offering reasoned and cogent rebuttals that successfully countered the tweets' anti-vaccine content.
However, this method  produced generic answers that fell short in addressing the particular concerns or issues raised in the tweet. These counter-arguments often lacked the contextual awareness and specificity needed to directly engage with and effectively refute the concerns expressed in the original tweets.

\vspace{2mm}
\noindent \textbf{(b) With Label Information:}
Numerous tweets addressed specific topics, including politics, pharmaceutical companies, side effects, and conspiracy theories. 
To enhance the effectiveness of the generated responses, we aimed to construct counter-arguments that explicitly addressed each key concern (or label) present in the tweets. To this end, we used the labels from the CAVES dataset—each of which comes with a short description, as shown in Table~\ref{tab:labels_map}. 
We then form a comprehensive instruction using the labels in a particular tweet, drawing inspiration from Chain-of-Thought (CoT)~\cite{wei2023chainofthoughtpromptingelicitsreasoning} prompting strategies. 
For tweets annotated with multiple labels, we concatenated the corresponding descriptions using the conjunction ``and''.
An example of such a prompt is: ``\texttt{Generate a strong counter-argument for the tweet: [text]. Discuss [Description 1] and [Description 2] $\cdots$ and [Description N]}.''
The counter-arguments generated using this enhanced prompt structure exhibited notable improvements in quality, demonstrating greater alignment with the tweet's underlying concerns and effectively addressing all specified labels.

\input{tables/dataset_exs}

\vspace{2mm}
\noindent \textbf{Sampling data:}
For our experiments, we prepared two datasets in total. In particular, we randomly selected 2,000 tweets from the CAVES dataset's 7,000 training examples and 990 tweets from its test set. From these, we constructed two parallel versions—one incorporating label descriptions and one without—which were then used to fine-tune various large language models (LLMs) for the counter-argument generation task.  Representative examples for both cases are provided in  Tables~\ref{tab:examples_without} and~\ref{tab:examples_with}.



%% file: tables/humanresponses.tex

\begin{table}[t]
\centering
\begin{tabular}{|l|c|c|c|c|c|}
\hline
\textbf{Model} & \textbf{BERT P} & \textbf{BERT R} & \textbf{BERT F1} \\
\hline
GPT-4o-mini & 0.82916 & 0.85242 & 0.84052 \\
\hline
Llama-1B  & 0.82473 & 0.85039 & 0.83725 \\
\hline
Phi-3  & 0.81893 & 0.84956 & 0.83387 \\
\hline
Tiny-Llama & 0.82920 & 0.85205 & 0.84036 \\
\hline
\end{tabular}
\caption{Comparison of Human and LLM Counter-Arguments Using ROUGE and BERT Scores calculated using human responses as gold standard}
\label{tab:human_llm_comparison}
\end{table}


%% file: tables/labeldesc.tex
\begin{table}[!t]
    \centering
    \small
    \begin{tabular}{|p{0.2\linewidth}|p{0.72\linewidth}|}
        \hline
        \textbf{Label} & \textbf{Description} \\
        \hline
        \textbf{Religious} & Religious beliefs and their influence on views about vaccines \\
        \hline
        \textbf{Political} & The political factors that affect perceptions of vaccine use \\
        \hline
        \textbf{Ingredients} & Concerns about the ingredients and chemical components in vaccines \\
        \hline
        \textbf{Unnecessary} & The importance and necessity of getting vaccinated to prevent diseases \\
        \hline
        \textbf{Conspiracy} & Conspiracy theories suggesting hidden motives behind vaccination efforts \\
        \hline
        \textbf{Mandatory} & The debate over personal choice versus mandates in vaccination policies \\
        \hline
        \textbf{Ineffective} & Evidence and reasons that support the effectiveness of vaccines \\
        \hline
        \textbf{Side-effect} & Potential side effects and adverse reactions associated with vaccines \\
        \hline
        \textbf{Pharma} & The role of pharmaceutical companies and concerns about profit motives \\
        \hline
        \textbf{Rushed} & Claims that vaccines were approved or developed without sufficient testing \\
        \hline
        \textbf{Country} & National biases and objections to vaccines produced by specific countries \\
        \hline
    \end{tabular}
    \caption{The different classes/labels in the CAVES dataset \cite{poddar2022caves} along with their descriptions}
    \label{tab:labels_map}
\end{table}

%% file: tables/dataset_exs.tex
\begin{table}[h!]
\centering
\renewcommand{\arraystretch}{1.2}
\begin{tabularx}{\columnwidth}{|X|}
\hline
\textbf{Tweet}: @Mike\_Pence @realDonaldTrump @pfizer The only way a ``vaccine'' could have been formulated in this amount of time is if the virus was man made to begin with. \\
\hline
\textbf{Prompt}: Generate a strong counter-argument for the tweet. \\
\hline
\textbf{Counter-Argument}: The assertion that a vaccine could only be developed quickly if the virus were man-made overlooks advancements in vaccine technology... The rapid development was possible due to years of prior research, global collaboration, and innovative platforms like mRNA. \\
\hline
\end{tabularx}
\caption{Example of a counter-argument generated without label information generated by GPT-4o-mini}
\label{tab:examples_without}
\end{table}

\begin{table}[h!]
\centering
\renewcommand{\arraystretch}{1.2}
\begin{tabularx}{\columnwidth}{|X|}
\hline
\textbf{Tweet}: @Mike\_Pence @realDonaldTrump @pfizer The only way a ``vaccine'' could have been formulated in this amount of time is if the virus was man made to begin with. \\
\hline
\textbf{Prompt}: Generate a strong counter-argument for the tweet. Talk about conspiracy theories suggesting hidden motives behind vaccination efforts and claims that vaccines were approved or developed without sufficient testing. \\
\hline
\textbf{Labels}: Conspiracy, rushed \\
\hline
\textbf{Counter-Argument}: While the creation of vaccines in a short timeframe can raise questions... the rapid development was enabled by global collaboration, prior research, and substantial funding... Conspiracy theories fail to consider rigorous clinical trials and FDA evaluations... Distrust fueled by such claims ultimately puts public health at risk.. \\
\hline
\end{tabularx}
\caption{Example of a counter-argument generated with label information by GPT-4o-mini}
\label{tab:examples_with}
\end{table}

%% file: sections/evaluation.tex
\section{Evaluation of the counter arguments}
To ensure the counter-arguments by GPT-4o-mini are impactful and relevant, we employ a comprehensive evaluation framework, combining human judgment and LLM-based assessment.

\subsection{Human Evaluation}
In this subsection we explain our methodology to evaluate our generated arguments (by GPT-4o-mini) by humans through a survey. 

\subsection{Data Selection for Survey}
For the survey, we sampled 100 data points from the dataset, each consisting of an anti-vaccine tweet paired with two corresponding counter-arguments generated by GPT-4o-mini: one without label descriptions and the other with label descriptions. Given the multi-label nature of the dataset, we employed stratified sampling to ensure balanced representation across all label categories. The final sample comprised 60 multi-labelled tweets and 40 single-labelled tweets.

\subsection{Survey Details}
To assess the effectiveness of counter-arguments generated by Large Language Models (LLMs) in addressing vaccine-related misinformation, we conducted a structured human evaluation survey. The study was hosted on the Prolific platform, which is widely used for collecting reliable and diverse human judgments. We recruited four participants—two who self-identified as vaccine-hesitant and two who were pro-vaccine—to ensure a balanced perspective. Participants were selected based on their consistently high approval ratings (99–100\%) and strong track record on the platform, which ensured data quality and reliability. Each participant was fairly compensated at a rate consistent with Prolific’s ethical pay guidelines.

\subsection{Survey Objective}
The primary objective of this survey was to assess whether incorporating label descriptions—which highlight key concerns (e.g., vaccine safety, efficacy, political motives)—into prompts improves the quality of counter-arguments generated by LLMs in response to anti-vaccine tweets.
\begin{itemize}
    \item \textbf{Prompt A} \textbf{(Baseline)}: The LLM was prompted using only the anti-vaccine tweet.
    \item  \textbf{Prompt B (Label Aware)}: The LLM was prompted with both the tweet and corresponding label descriptions that explain the main objections raised.
\end{itemize}
We conducted a two-phase user study to evaluate the impact of incorporating explicit contextual information—specifically, label descriptions—on the quality of counter-arguments generated by Large Language Models (LLMs). We first conducted a preliminary pilot study on 20 tweets to identify and resolve potential issues related to survey design, question framing, and interface usability. Following this, a main evaluation was carried out on 80 anti-vaccine tweets.

Each participant was presented with 100 tweets, each accompanied by the tweet text, a set of assigned labels (capturing key concerns), and their corresponding descriptions. For each tweet, participants were shown two counter-arguments: one generated using a baseline prompt (Prompt A) without label descriptions, and another generated using an enhanced prompt (Prompt B) that included the label descriptions. To mitigate positional bias, the order of the two responses was randomized.

Participants were asked to choose the counter-argument they found more effective in addressing the concerns raised by the tweet and to briefly justify their selection. Evaluation criteria included comprehensiveness (i.e., whether all key concerns were addressed), clarity, factual accuracy, and overall persuasiveness. To ensure traceability and accurate analysis, structured metadata was embedded in the Google Forms used for data collection. This metadata enabled us to de-randomize the responses during post-survey analysis and accurately attribute each counter-argument to its corresponding prompting strategy.

\input{sections/evaluationresults}

%% file: sections/evaluationresults.tex
\subsection{Human Evaluation Results}


\input{tables/human_survey}
To assess the quality of generated counter arguments, we conducted a human evaluation study on \textbf{100 }tweets. We recruited 4 annotators—two pro-vaccine and two anti-vaccine—where each annotator was asked to choose between two counter-arguments (A and B) for each tweet, and we determined the final vote for each tweet by majority voting.

Since each tweet in our study was evaluated by four annotators, we represented the outcomes using consistent vote ratio bins (e.g., 3:1, 2:2, etc.). The distribution of Counter-Argument B:A vote ratios across all 100 tweets is shown in Table~\ref{tab:vote_ratio_combined}.

From this combined evaluation, we observe that 33 tweets had a 3:1 vote ratio in favor of Counter-Argument B, making it the most common outcome. In addition, 26 tweets were evenly split (2:2), and 24 tweets had a 1:3 ratio, favoring Argument A. On the extremes, 15 tweets had unanimous votes for Argument B (4:0), while only 3 tweets had unanimous votes for Argument A (0:4). These results clearly indicate a strong preference for Counter-Argument B, with over \textbf{75\%} of tweets either favoring B or showing equal support.


This pattern is consistent throughout the dataset, indicating that annotators, irrespective of their position on vaccination, respond more favorably to B-style counter-arguments that are constructed using label-aware prompting. The prevalence of 3:1 and 4:0 ratios favoring B suggests that these answers were frequently chosen over others, rather than just being slightly superior. Cases where both arguments were equally compelling or where stylistic preferences led to a split are highlighted by the noteworthy number of 2:2 ties. As demonstrated by the consistent human preferences for Argument B, these results thus lend credence to the idea that adding label descriptions to the creation of counterarguments produces more convincing and well-received outputs.\\
\vspace{-0.270cm}
\subsection{LLM Evaluation}
We use GPT-4o to evaluate the same set of tweets shown to human annotators. The model is prompted appropriately and run four times per tweet, allowing us to derive a majority vote from its multiple responses.
Out of all evaluations conducted across 100 anti-vaccine tweets, \textbf{24.5\% of votes favored Counter Argument A} (generated without label descriptions), while \textbf{55.5\% favored Counter Argument B} (generated using label descriptions). The remaining \textbf{20\% of the responses indicated equal votes}, with evaluators rating both counter-arguments as equally effective. This distribution suggests that providing explicit label-based context in the prompt improved the perceived quality and effectiveness of the generated responses. 
\vspace{-0.15cm}

\input{tables/agreement}
\subsection{Analysis on Human Judgement}
We answer four key questions to analyze human preferences Representative examples for parts 1 and 2 are shown in Appendix~\ref{appendix:human_judgment}


\begin{enumerate}
    \item \textbf{Why was Counter-Argument A favored?} \\
Counter-Argument A was frequently favored when it offered concise, fact-based refutations that specifically addressed the issues raised in the initial tweet, particularly in cases where the associated labels did not strongly correlate with the tweet’s core content.  These responses tended to be succinct, outlining their points in an organized and effective way without needless elaboration. They were also measured and professional in tone, using reasoned arguments rather than sentimental or dramatic language. All of these qualities combined to give human raters a greater sense of persuasiveness and credibility. Table \ref{table: A better than B} shows an example where the counter-argument A was preferred than counter-argument B.

    \item \textbf{Why was Counter-Argument B favored?} \\
    Counter-Argument B were favoured when they were more sympathetic and approachable. They were felt more relatable and thoughtful because they frequently acknowledged the emotional factors such as fear, personal preference, or mistrust that underlie vaccine hesitancy. They talked about the specific labeled concerns in the tweet (eg. ingredients, side-effects..) which makes them more targeted and relevant compared to generic responses given by A especially when the tweets were multi-labeled. Argument B also tended to be concise and conversational, avoiding unnecessary technical detail while still maintaining factual accuracy. By striking a balance between emotional awareness, factual grounding, and alignment with tweet-specific concerns, these responses were consistently judged as more persuasive and effective. Table \ref{Table: B was better than A} shows an example where the counter-argument B was preferred over counter-argument A. 
    \item \textbf{When did both arguments agree?} \\
    Human raters tended to show strong agreement when one of the counter-arguments clearly addressed the core concern raised in the anti-vaccine tweet. When the tone of the argument aligned with the type of false information, consensus was especially high. For example, direct, logical reasoning was the most effective way to refute conspiracy-laden claims. Furthermore, they consistently preferred arguments that grounded their answers in specific facts, like citing research studies or describing the workings of public health systems. These characteristics increased the likelihood of agreement among raters by fostering a common sense of legitimacy and applicability.
    \item \textbf{When did both arguments disagree?} \\
    When both counterarguments were well-written but used different rhetorical stances one relying on facts and logic, the other on emotional appeal differences between raters frequently emerged. Different rater's interpretations of the original tweet also contributed; for example, some raters thought a post emphasized personal freedom, while others raised concerns about vaccine safety. Individual stylistic preferences also played a role in these situations; some raters preferred a tone that was calm and neutral, while others were more receptive to arguments that expressed emotional resonance or empathy. Table \ref{Table: Comparision between A and B} shows a comparison between counter-argument A and counter-argument B. 
\end{enumerate}

%% file: tables/human_survey.tex
\begin{table}[h]
\centering
\begin{tabular}{|c|c|}
\hline
\textbf{B:A Vote Ratio} & \textbf{Number of Tweets} \\
\hline
0:4 & 3 \\
1:3 & 24 \\
2:2 & 26 \\
3:1 & 33 \\
4:0 & 15 \\
\hline
\end{tabular}
\caption{Distribution of B:A vote ratios across 100 tweets after annotated by 4 annotators. This table compares the results of baseline (Counter Argument A) and label-aware prompts (Counter Argument B)}
\label{tab:vote_ratio_combined}
\end{table}

%% file: tables/agreement.tex

%% file: sections/twostepcot.tex
\begin{figure}[h]
  \centering
  \includegraphics[trim=0 65 0 65, clip, width=0.95\linewidth]{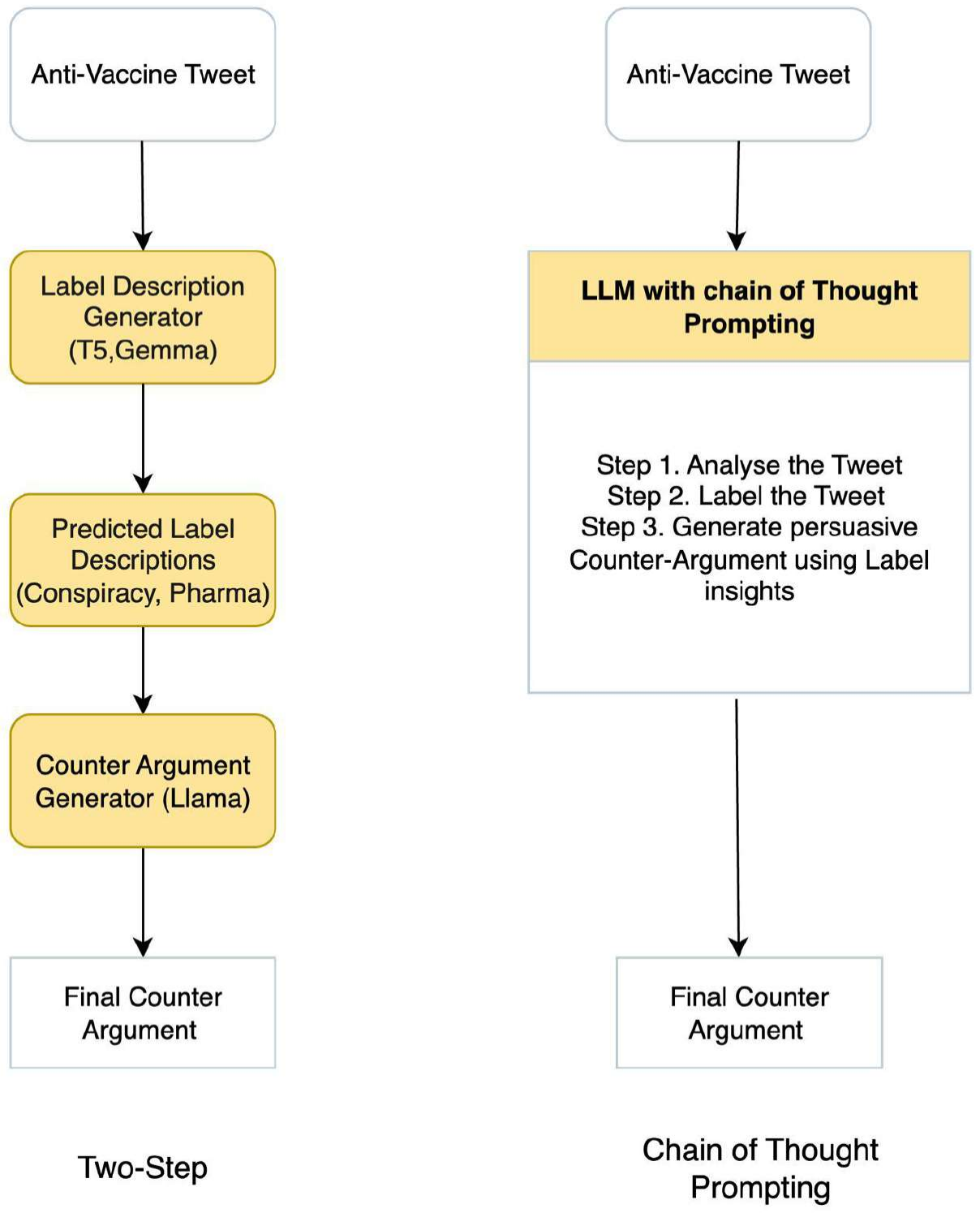}
  \captionsetup{width=0.9\linewidth} 
  \caption{
    Overview of the proposed approaches for generating counter-arguments to anti-vaccine tweets using label descriptions. The Two-Step approach involves predicting label descriptions using a separate model and then generating counter-arguments. The Chain-of-Thought approach integrates reasoning and label inference within a single model prompt to generate persuasive counter-arguments.
  }
  \label{fig:architecture_pipeline}
\end{figure}

\subsection{Motivation}
In real-world social media environments, tweets containing misinformation rarely come with predefined labels or explanations of underlying concerns, making it challenging for automated systems to generate precise counter-arguments. Relying solely on such labels limits the practicality and scalability of counter-misinformation efforts. To address this, we introduce a two-step pipeline that mimics human reasoning: first, identifying the core concerns embedded in the tweet, and then generating a contextually grounded counter-argument. This approach not only eliminates dependency on manually annotated labels but also improves robustness in unseen scenarios. Furthermore, by incorporating Chain-of-Thought prompting, we encourage the model to explicitly reason through the content of the tweet before crafting a response. This structured reasoning improves interpretability and leads to more nuanced, targeted rebuttals—critical for effectively addressing complex and multi-faceted vaccine skepticism.

\subsection{Two-Step Approach}
\sloppy
In this approach, we adopt a two-step methodology that leverages three categories of language models: encoder–decoder models such as \textbf{T5-Base} and \textbf{T5-Large} \cite{raffel2023exploringlimitstransferlearning}, decoder-only models \textbf{Gemma-2B} and \textbf{LLaMA-3B} \cite{gemmateam2024gemma2improvingopen,grattafiori2024llama3herdmodels}, and encoder-only models \textbf{RoBERTa-Large} and \textbf{Covid-Twitter-Bert} \cite{liu2019robertarobustlyoptimizedbert,müller2020covidtwitterbertnaturallanguageprocessing}.
\fussy

\smallskip
\noindent\textbf{Generating Label Descriptions.} 
Following \cite{Poddar_Mukherjee_Khatuya_Ganguly_Ghosh_2024}, we use encoder–decoder models to generate label descriptions for anti-vaccine tweets. 
We instruction fine-tune these models on 7{,}000 anti-vaccine tweets paired with their corresponding label descriptions from the CAVES train split and evaluate them on 2{,}000 tweets from the test split. 
We also instruction fine-tune the decoder-only models using the same setup.

\smallskip
\noindent\textbf{Matching Label Descriptions.} 
\sloppy
Since the generated label descriptions may not exactly match the gold standard in the CAVES dataset, we adopt the mapping strategy proposed by \cite{Poddar_Mukherjee_Khatuya_Ganguly_Ghosh_2024}. 
We split each generated description into sentences (if multiple), encode both generated and ground-truth descriptions using a pre-trained Sentence-BERT encoder \cite{reimers2019sentencebertsentenceembeddingsusing}, compute cosine similarity scores with all available ground-truth label descriptions, and assign the label corresponding to the highest similarity. 
This ensures that the final assigned labels are semantically consistent and aligned with the CAVES taxonomy.
\fussy

\smallskip
\noindent\textbf{Multi-Label Classification.} 
We fine-tune the encoder-only models (\textbf{RoBERTa-Large}, \textbf{Covid-Twitter-Bert}) on an 11-label multi-label classification task using the same dataset. 
This enables them to predict label categories directly without explicit prompt-based instruction tuning. 
These models can infer descriptive label phrases or raw identifiers, making them suitable for downstream tasks such as counter-argument generation when explicit label metadata is unavailable.

\smallskip
\noindent\textbf{Metrics.} 
We evaluate the models using three standard metrics. 
Given the predicted and gold standard label sets for each tweet, we compute the F1-score for each of the 11 classes (as described in Table~\ref{tab:labels_map}) and report \textbf{(i)} Macro-average F1 score, \textbf{(ii)} Micro-average F1 score, and \textbf{(iii)} Average accuracy across labels. Results are shown in Table \ref{tab:label_prediction_results}.
All metrics are calculated using the \textsc{scikit-learn} package \cite{pedregosa2011scikit}.

\medskip
In the second step, we use the predicted labels as input to small language models (SLMs) such as Gemma, LLaMA, and Phi-3 Mini to generate contextually grounded counter-arguments. 
The evaluation of these counter-arguments is discussed in subsequent sections.

\input{tables/labelresults}
\subsection{COT Approach }
Here, we perform few-shot and zero-shot prompting on \textbf{Gemma 9B} and \textbf{LLaMA 8B }models by carefully designing complex \textbf{Chain-of-Thought (CoT) prompts} as shown in Table \ref{tab: Two-step and COT} \cite{wei2023chainofthoughtpromptingelicitsreasoning}. These prompts guide the models through a structured reasoning process by first asking them to analyze the given tweet, identify underlying concerns or themes, and then generate the most appropriate label descriptions. This intermediate reasoning step ensures that the models develop a contextual understanding before producing label outputs, thereby improving the relevance and coherence of their predictions.

\input{tables/prompts}

\subsection{Knowledge Distillation and Model Training}
Recent advancements in large language models (LLMs), such as GPT-4 and LLama, have demonstrated remarkable performance across a wide spectrum of natural language processing tasks. However, deploying these models in real-world applications particularly within resource constrained environments remains challenging due to their substantial computational and memory demands.

To mitigate these limitations, we adopt a knowledge distillation framework \cite{xu2024surveyknowledgedistillationlarge}, where in a compact \textit{student} model is trained to emulate the behavior of a larger, pre-trained \textit{teacher} model. This strategy allows us to preserve much of the performance of the larger model while significantly reducing inference latency and memory footprint, making the deployment of LLMs more practical and scalable in constrained settings.
\\
Applying knowledge distillation to small language models (SLMs) such as Gemma, Phi-3, and LLaMA offers several practical advantages that enhance their real-world utility. It improves computational efficiency by enabling faster inference and reducing resource usage, which makes these models suitable for deployment in resource-constrained environments such as edge devices. The technique also supports task-specific adaptation by allowing models to be fine-tuned on domain relevant tasks using fewer resources. Through soft-label supervision and representation alignment, knowledge distillation helps smaller models retain much of the reasoning ability and accuracy of larger counterparts. This makes them capable of delivering strong performance without the high computational cost. It also enables scalability, making it feasible to deploy language models in settings that demand real-time responses or operate under limited hardware constraints, such as mobile platforms or IoT systems. 
Thus, knowledge distillation bridges the gap between state-of-the-art model performance and practical deployment requirements, making it a highly effective strategy for enhancing the utility of models like Gemma and LLaMA.

\subsection{Training Setup}

We fine-tuned a set of Small Language Models (SLMs), including \textbf{LLaMA-3.2-3B}, \textbf{Gemma-2B}, and \textbf{Phi-3 Mini}, using knowledge distillation techniques for the task of counter-argument generation. 
Our training pipeline was implemented using the \texttt{transformers} and \texttt{peft} libraries \cite{xu2023parameterefficientfinetuningmethodspretrained}, in combination with \texttt{bitsandbytes} for 4-bit quantization and the \texttt{SFTTrainer} from the \texttt{trl} library for supervised fine-tuning. 
All training was performed on GPUs with support for \texttt{bfloat16} and \texttt{Flash Attention 2} when available, defaulting to \texttt{float16} and \texttt{SDPA} otherwise.

\smallskip
\noindent\textbf{Quantization and LoRA Configuration.}  
We used 4-bit quantization with the NF4 format and enabled double quantization to reduce memory footprint and speed up training. 
To further optimize resource efficiency, we applied Low-Rank Adaptation (LoRA) \cite{hu2021loralowrankadaptationlarge} with a rank of 4, $\alpha = 32$ and a dropout of 0.1. 
LoRA modules were injected into key attention and feedforward projection layers. 
We also enabled gradient checkpointing to minimize activation storage during training.

\medskip
\noindent\textbf{Optimization Details.}  
We used the AdamW optimizer with a learning rate of \( 1 \times 10^{-4} \), a linear learning rate scheduler, and a warmup ratio of 10\%. 
The models were trained for 2 epochs using a per-device batch size of 2 and gradient accumulation steps of 2, effectively simulating a batch size of 4. The maximum input sequence length was set to 512 tokens. We evaluated model performance every 100 steps and at the end of each epoch. All models were trained with fp16 mixed precision.
We evaluated model performance every 100 steps and at the end of each epoch. 
All models were trained using \texttt{fp16} mixed precision.

\medskip
\noindent\textbf{Prompt Formats and Data.}  
We use the same prompts described in Section 3.2 for the Dataset Preparation using two prompt formats: (i) zero-shot prompts without label descriptions, and (ii) zero-shot prompts that included descriptive labels. For each format, we used 2000 data points for training and 990 for evaluation. The training data was structured in a \textit{ChatML-style} format using the \textit{DataCollatorForCompletionOnlyLM}, with clearly separated user and assistant segments to improve instruction-following behavior.

%% file: tables/labelresults.tex
\begin{table}[H]
\small
\centering
\begin{tabular}{|l|c|c|c|}
\hline
\textbf{Model} & \textbf{F1-Macro} & \textbf{F1-Micro} & \textbf{Accuracy} \\
\hline
T5-Base             & 0.876  & 0.903  & 0.796  \\
T5-Large            & 0.882  & 0.892  & 0.760  \\
Gemma-2B            & 0.511  & 0.585  & 0.537  \\
LLaMA-3B            & 0.497  & 0.567  & 0.545  \\
RoBERTa-Large       & 0.6783 & 0.7408 & 0.5964 \\
COVID-Twitter-BERT  & 0.6785 & 0.7446 & 0.6146 \\
\hline
\end{tabular}
\caption{Performance of Label Prediction Models using Sentence-BERT Similarity.}
\label{tab:label_prediction_results}
\end{table}

%% file: tables/prompts.tex
\begin{table*}[t]
\small
\centering

\label{tab:prompt_examples}
\begin{tabularx}{\textwidth}{|X|}
\hline
\textbf{Prompt for Label Prediction} \\ \hline
\ttfamily
Instruction: First read the task description. There could be multiple category descriptions for a tweet. \newline
Task: Multi-label Text Classification \newline
Description: Generate label description for the given texts. \newline
Tweet: @sandraburgess3 They have no idea, they can't control the Flu with vaccine so what makes them think they can control covid 19 because all flu seems to have a variant of covid in it as do most infectious diseases including Ebola!!! \\
\hline
\textbf{Zero-Shot Prompt} \\ \hline
\ttfamily
You are an expert in analyzing social media content and generating counter-arguments to misinformation. Follow these steps: \newline
1. \textbf{Step 1:} Analyze the main themes or issues discussed in the tweet. Based on your analysis, assign relevant labels that describe the misinformation topics involved. Choose the most appropriate labels from the following list: [Label 1 , Label 2, ...] \newline
2. \textbf{Step 2:} Using the labels and the content of the tweet, generate a clear, evidence-based counter-argument that addresses the claims in the tweet. \newline
Tweet: \{tweet\} \newline
Your response: \newline
1. Labels: \newline
2. Counter-argument: \\
\hline
\textbf{Few-Shot Prompt} \\ \hline
\ttfamily
You are an expert in analyzing social media content and generating counter-arguments to misinformation. Follow these steps: \newline
1. \textbf{Step 1:} Analyze the main themes or issues discussed in the tweet. Based on your analysis, assign relevant labels that describe the misinformation topics involved. Choose the most appropriate labels from the following list: [Label 1 , Label 2, ...] \newline
2. \textbf{Step 2:} Using the insights from the labels, craft a well-reasoned and persuasive counter-argument that respectfully addresses the concerns raised in the tweet, corrects misinformation, and promotes scientific understanding. \newline
\textbf{Example 1:} \newline
Tweet: "\{example tweet\}" \newline
1. Labels - [Label1, Label2, ...] \newline
2. Counter-Argument - \{Counter-Argument\} \newline
\textbf{Your Response:} \newline
Tweet: \{tweet\} \newline
1. Labels: \newline
2. Counter-argument: \\
\hline
\end{tabularx}
\caption{Examples of Prompts Used for Label Description Prediction and Counter-Argument Generation using COT}
\label{tab: Two-step and COT}
\end{table*}

%% file: sections/knowledgedistill.tex
\section{Experiments and Results}
To evaluate the effectiveness of incorporating label descriptions during counter-argument generation, 
we utilize the  \counterdata{} introduced in Section~\ref{sec:Generating Counter Arguments}. 
The \counterdata{} dataset comprises two distinct subsets constructed using \textbf{GPT-4o-mini} through zero-shot prompting. 
The first subset, termed \textit{NoLabelPromptSet}, is generated using a basic zero-shot prompt without any label annotations 
and contains \textbf{2,000 training examples}. 
The second subset, termed \textit{LabelAwarePromptSet}, is generated using a zero-shot prompt that includes 
\textbf{descriptive labels} associated with the misinformation present in each tweet, 
enabling label-aware counter-argument generation. 
This subset contains \textbf{2,000 training examples} and \textbf{990 evaluation examples}, 
which serve as the held-out test set for all experiments.

For evaluation, we take \textbf{990 evaluation points} from the \textit{LabelAwarePromptSet} as our test set. We conduct three sets of experiments to assess whether the inclusion of label descriptions leads to improved model performance. Specifically, we fine-tune and evaluate the following small language models: Gemma-2B, LLaMA-3B, LLaMA-1B, Phi-3-Mini, and TinyLLaMA. We conduct 3 experiments described below:
\begin{itemize}
    \item The model is fine-tuned on \textit{NoLabelPromptSet}, where the prompts do not include any label descriptions. During evaluation on the \textit{LabelAwarePromptSet} test set, label descriptions are also not provided.
    
    \item The model is fine-tuned on \textit{NoLabelPromptSet}, but the prompts are augmented with all label descriptions. During evaluation on the \textit{LabelAwarePromptSet} test set, the label descriptions are also included in the prompts..

    \item The model is fine-tuned on \textit{LabelAwarePromptSet}, using prompts that  include label descriptions. During evaluation on the \textit{LabelAwarePromptSet} test set, the label descriptions are also included in the prompts
\end{itemize}

\subsection{Evaluation}

We evaluate all five models across the three experimental settings using standard text generation metrics: \textbf{ROUGE-2}, \textbf{ROUGE-L}, and \textbf{BERT-based Precision}, \textbf{Recall}, and \textbf{F1} scores. 
These metrics are computed by comparing the model-generated counter-arguments with the corresponding references from the test set.

\medskip
\input{tables/evaluationresults}

\medskip
\noindent\textbf{1. ROUGE-2 Score.} \\
ROUGE-2 evaluates the bigram overlap between generated and reference text:
\[
\text{ROUGE-2} = 
\frac{\sum_{n=2}^{N} \text{count}_{\text{match}}(n)}{\sum_{n=2}^{N} \text{count}_{\text{reference}}(n)}.
\]

\medskip
\noindent\textbf{2. ROUGE-L Score.} \\
ROUGE-L measures the longest common subsequence (LCS)-based F-measure:
\[
\text{ROUGE-L} = 
\frac{F_{1}}{F_{1} + \lambda \cdot \text{missed}},
\]
where \( F_{1} \) is the F-measure and \( \lambda \) controls the importance of recall relative to precision.

\medskip
\noindent\textbf{3. BERT Precision.} \\
BERT Precision computes the average cosine similarity between each token embedding in the generated text and its best match in the reference text:
\begin{equation}
\text{BERT Precision} = 
\frac{1}{|T|} \sum_{t \in T} \max_{r \in R} \text{CosineSim}(E_t, E_r),
\end{equation}
where \( T \) is the set of generated tokens and \( R \) is the set of reference tokens.

\medskip
\noindent\textbf{4. BERT Recall.} \\
Similarly, BERT Recall measures how well the reference tokens are covered by the generated tokens:
\begin{equation}
\text{BERT Recall} = 
\frac{1}{|R|} \sum_{r \in R} \max_{t \in T} \text{CosineSim}(E_r, E_t).
\end{equation}

\medskip
\noindent\textbf{5. BERT F1.} \\
Finally, BERT F1 is computed as the harmonic mean of BERT Precision and Recall:
\begin{equation}
\text{BERT F1} = 
2 \times \frac{\text{BERT Precision} \times \text{BERT Recall}}
{\text{BERT Precision} + \text{BERT Recall}}.
\end{equation}

\subsection{Analysis}
As seen in Tables 8, 9, and 10, the ROUGE and BERT scores consistently improve from Experiment 1 to Experiment 3. This trend supports our claim that incorporating and training with label descriptions enhances the quality and coherence of the generated counter-arguments.

\input{tables/cotandtwostep}
\subsection{Model Evaluation Results for Two Step Approach and COT Reasoning }


The results in Table 11 compare the performance of the Two-Step and Chain-of-Thought (CoT) reasoning approaches for the Gemma and Llama 3B models.For Two-Step we use the labels generated by T5-Base model and then feed these in the decoder models. The analysis indicates that the Two-Step approach consistently outperforms CoT reasoning across both models. Specifically, Two-Step surpasses the performance of Experiment 2 and closely approaches the results of Experiment 3, demonstrating its effectiveness in enhancing argument quality and coherence. In contrast, CoT reasoning yields results slightly below those of Experiment 3 and marginally better than Experiment 2. Experiment 1 continues to show the weakest performance overall, reaffirming the impact of incorporating structured reasoning and label descriptions.

%% file: tables/evaluationresults.tex
\begin{table*}[t]
\centering
\begin{tabular}{|l|c|c|c|c|c|}
\hline
\textbf{Model} & \textbf{ROUGE-2} & \textbf{ROUGE-L} & \textbf{BERT P} & \textbf{BERT R} & \textbf{BERT F1} \\
\hline
Gemma-2B    & 15.73  & 24.52  & 0.8791 & 0.8944 & 0.8865 \\
LLaMA-3.2-3B    & 11.55  & 22.61  & 0.8704 & 0.8931 & 0.8816 \\
LLaMA-3.2-1B    & 9.03   & 20.54  & 0.8666 & 0.8871 & 0.8766 \\
Phi-3‑Mini      & 4.50   & 14.28  & 0.8477 & 0.8724 & 0.8598 \\
TinyLLaMA       & 10.24  & 21.86  & 0.8698 & 0.8824 & 0.8761 \\
\hline
\end{tabular}
\caption{Experiment 1: Models fine-tuned on \textit{NoLabelPromptSet} without label descriptions. Evaluation on the \textit{LabelAwarePromptSet }test set also excludes label descriptions.}
\label{tab:exp1_results}
\end{table*}

\begin{table*}[t]
\centering
\begin{tabular}{|l|c|c|c|c|c|}
\hline
\textbf{Model} & \textbf{ROUGE-2} & \textbf{ROUGE-L} & \textbf{BERT P} & \textbf{BERT R} & \textbf{BERT F1} \\
\hline
Gemma-2B          & 16.98  & 27.51  & 0.8777 & 0.8938 & 0.8851 \\
LLaMA-3.2-3B   & 13.04  & 24.54  & 0.8752 & 0.8953 & 0.8851 \\
LLaMA-3.2-1B   & 10.59  & 21.80  & 0.8723 & 0.8867 & 0.8794 \\
Phi-3‑Mini     & 5.59   & 15.10  & 0.8509 & 0.8757 & 0.8631 \\
Tiny-LLaMA     & 9.76   & 22.70  & 0.8702 & 0.8826 & 0.8763 \\
\hline
\end{tabular}
\caption{Experiment 2: Models fine-tuned on \textit{NoLabelPromptSet} with label descriptions included in prompts. Evaluation on \textit{LabelAwarePromptSet} test set also includes label descriptions.}
\label{tab:exp2_results}
\end{table*}

\begin{table*}[t]
\centering
\begin{tabular}{|l|c|c|c|c|c|}
\hline
\textbf{Model} & \textbf{ROUGE-2} & \textbf{ROUGE-L} & \textbf{BERT P} & \textbf{BERT R} & \textbf{BERT F1} \\
\hline
Gemma-2B          & 20.76 & 30.49 & 0.9000 & 0.8890 & 0.8950 \\
LLaMA-3.2-3B      & 15.03 & 25.34 & 0.8930 & 0.8857 & 0.8894 \\
LLaMA-3.2-1B      & 14.06 & 24.52 & 0.8660 & 0.8870 & 0.8760 \\
Phi-3-Mini        &  8.09 & 17.85 & 0.8690 & 0.8660 & 0.8670 \\
Tiny-LLaMA        & 12.54 & 22.97 & 0.8910 & 0.8720 & 0.8820 \\
\hline
\end{tabular}
\caption{Experiment 3: Models fine-tuned on \textit{LabelAwarePromptSet} using prompts that include label descriptions. Evaluation
on \textit{LabelAwarePromptSet} test set also includes label descriptions.}
\label{tab:exp3_results}
\end{table*}


%% file: tables/cotandtwostep.tex
\begin{table*}[t]
\centering

\label{tab:reasoning_results}
\begin{tabular}{|l|l|c|c|c|c|c|}
\hline
\textbf{Model} & \textbf{Approach} & \textbf{ROUGE-2} & \textbf{ROUGE-L} & \textbf{BERT P} & \textbf{BERT R} & \textbf{BERT F1} \\
\hline
Gemma-9B         & Two-Step Reasoning  & 19.35 & 28.14 & 0.898 & 0.887 & 0.893 \\ \cline{2-7}
                 & CoT Reasoning       & 15.45 & 26.28 & 0.853 & 0.871 & 0.887 \\ \hline
LLaMA-3.2-8B      & Two-Step Reasoning  & 14.13 & 22.44 & 0.888 & 0.882 & 0.885 \\ \cline{2-7}
                 & CoT Reasoning       & 12.12 & 21.04 & 0.863 & 0.881 & 0.878 \\
\hline
\end{tabular}
\caption{Results for Two-Step vs. Chain-of-Thought (CoT) Reasoning Approaches for Gemma and LLaMA-3 Models}
\end{table*}

%% file: sections/conclusion.tex
\section{Conclusion}
This research addresses the challenge of generating effective counter-arguments to anti-vaccine tweets using both Large Language Models (LLMs) and Small Language Models (SLMs). To facilitate this task, we first curated a high-quality dataset by prompting GPT-4o Turbo to generate counter-arguments for real-world anti-vaccine tweets, which were then evaluated and refined through human review for clarity, accuracy, and persuasiveness. This dataset served as the foundation for all subsequent experiments

We conducted three main experiments to assess how different prompt configurations affect the quality of generated counter-arguments. In Experiment 1, models were prompted with the raw tweet alone, resulting in lower coherence and weak performance across standard metrics. Experiment 2 added stance labels, which slightly improved contextual understanding. The most notable gains were observed in Experiment 3, where label descriptions were included in the prompt. This addition consistently led to higher scores in  metrics, confirming that detailed contextual cues enable models to generate more precise and persuasive counter-arguments.

Further, we evaluated reasoning strategies by comparing Two-Step prompting and Chain-of-Thought (CoT) reasoning on Gemma and LLaMA 3.2 3B models. Both strategies improved performance over the baseline, with the Two-Step approach slightly outperforming CoT in most metrics. However, neither surpassed the results of Experiment 3, reinforcing the effectiveness of descriptive prompting over pure reasoning techniques in this context.

Our research demonstrates that prompt engineering especially through the use of descriptive label information and reasoning strategies can significantly influence the quality of counter-arguments generated by LLMs and SLMs. It also highlights that smaller models, when guided effectively, can generate credible and coherent arguments, thereby offering a cost-effective alternative to larger models for specific tasks. \\
Overall, this research provides a foundational framework for enhancing automated argumentation, contributing to the broader goal of combating misinformation with informed, diverse, and accurate counterpoints.

%% file: sections/appendix.tex
\section*{Appendix} \label{appendix:examples}
\addcontentsline{toc}{section}{Appendix}

%






\section{Examples of Human Judgment Cases}
\label{appendix:human_judgment}

In this section, we provide illustrative examples corresponding to our human evaluation analysis (Section~4.8). Each example shows the original tweet, the prompt, the generated counter-arguments (A and B), and indicates which case it exemplifies: why Counter-Argument A was favored, why Counter-Argument B was favored, or when both were rated equally strong.


\begin{table}[h!]
\centering
\caption{Case 1: Example where Counter-Argument A was preferred}
\renewcommand{\arraystretch}{1.2}
\begin{tabularx}{\columnwidth}{|X|}
\hline
\textbf{Tweet:} Vaccine would be against my religion. Which is the 91st Psalm is the best protection there is. \\ \hline

\textbf{Counter-Argument A:} While personal beliefs and religious convictions are important, the overwhelming scientific consensus supports the efficacy of vaccines in preventing disease and protecting public health. Vaccines have been crucial in controlling outbreaks of diseases that once caused widespread illness and death, such as measles, polio, and influenza. The 91st Psalm, while providing spiritual comfort and strength, does not replace the need for medical interventions that save lives. Additionally, many religious leaders and organizations endorse vaccination as a means of loving one’s neighbor and protecting the vulnerable. \\ \hline

\textbf{Counter-Argument B:} While it's understandable that religious beliefs can significantly influence personal health decisions, public health data consistently shows that vaccines are vital in preventing serious illnesses, many of which can be fatal. The 91st Psalm emphasizes faith and trust in divine protection; however, it should not be seen as a substitute for scientifically proven medical interventions. Vaccines not only protect the individual but also contribute to community immunity, safeguarding vulnerable populations who cannot be vaccinated for medical reasons. Religious exemptions to vaccination have raised public health concerns, as illustrated in various studies and articles on vaccine efficacy and the importance of herd immunity. \\ \hline

\textbf{Human Judgment:} Annotators preferred Counter-Argument A because it addressed the original tweet's religious framing head-on while maintaining a strong scientific position. Instead of dismissing it outright, the response respected the religious foundation by mentioning Psalm 91. By linking faith and science and referencing the support of religious leaders and organizations, Argument A enhanced its persuasiveness. In contrast, Argument B leaned more heavily on technical explanations and public health data, making it feel less personally relevant to the tweet’s concerns. \\ \hline
\end{tabularx}
\label{table: A better than B}
\end{table}

\begin{table}[h!]
\centering
\caption{Case 2: Example where Counter-Argument B was preferred}
\renewcommand{\arraystretch}{1.2}
\begin{tabularx}{\columnwidth}{|X|}
\hline
\textbf{Tweet:} @iamErica\_Mena what ’ s happening ? Does it have to do with that mRNA protein - idk, but it ’ s doing something that ’ s causing a chemical reaction that kills . So I ’ m arguing with a veterinarian about the vaccine  \\ \hline
\textbf{Labels: }ingredients side-affect \\ \hline

\textbf{Counter-Argument A:} While concerns about vaccines, particularly mRNA technology, are common, it's essential to understand that these vaccines have undergone rigorous testing for safety and efficacy. The mRNA vaccines, such as those developed by Pfizer and Moderna, instruct cells to produce a harmless piece of the spike protein found on the surface of the SARS-CoV-2 virus, which helps our immune system recognize and fight the actual virus. Studies have consistently shown that the benefits of vaccination far outweigh the risks, significantly reducing hospitalizations and deaths (CDC, 2021). Furthermore, trusted sources like the World Health Organization and the Centers for Disease Control and Prevention provide evidence-based information that underscores the importance of vaccines in managing infectious diseases.\\ \hline

\textbf{Counter-Argument B:} While concerns about vaccines, including their ingredients and potential side effects, are understandable, it’s crucial to examine the overwhelming scientific consensus on their safety and efficacy. Vaccines undergo rigorous testing in clinical trials, and regulatory bodies like the CDC and WHO continuously monitor their safety. The mRNA technology used in vaccines has been shown to be safe and effective in preventing serious diseases, including COVID-19. The likelihood of severe adverse reactions is extremely low compared to the risks posed by the diseases they prevent. The benefits of vaccination—such as community immunity and the prevention of outbreaks—vastly outweigh the risks. \\ \hline

\textbf{Human Judgment:} Counter-Argument B was favored because it directly and concisely addressed the tweet’s concerns about ingredients and side effects without overwhelming the reader with technical details. Annotators felt it provided clear reassurance about mRNA safety, highlighted the low risks of adverse effects, and remained focused on the specific issues raised in the tweet. In contrast, Counter-Argument A, though informative, was seen as overly technical and less relatable, making B the more effective and persuasive response. \\ \hline
\end{tabularx}
\label{Table: B was better than A}
\end{table}

\begin{table*}[h!]
\centering
\renewcommand{\arraystretch}{1.3}
\begin{tabularx}{\textwidth}{|X|}
\hline
\textbf{Tweet}: \text{@RepSwalwell And yet you want a Nationwide vaccine so that people can get injected with it} \\ \hline

\textbf{Label}: Mandatory \\ \hline

\textbf{Counter-Argument A}: While concerns about vaccines are valid and deserve examination, the overwhelming scientific consensus supports their safety and efficacy in preventing serious diseases. Vaccines have played a crucial role in reducing the incidence of illnesses like measles, polio, and more recently, COVID-19. The widespread adoption of vaccines not only protects individuals but also helps build community immunity, safeguarding those who cannot be vaccinated due to medical reasons. Extensive research, including studies published by organizations like the CDC and WHO, shows that vaccines undergo rigorous testing for safety before approval. \\ \hline

\textbf{Counter-Argument B}: While personal choice is an important aspect of healthcare, the argument for mandatory vaccinations is fundamentally about public health and safety. Vaccines have been proven to reduce the spread of infectious diseases, protecting not only the individuals who receive them but also those who cannot be vaccinated due to medical conditions. Mandates can lead to higher vaccination rates, which in turn can help achieve herd immunity and prevent outbreaks. The CDC outlines that vaccines save millions of lives each year and prevent the spread of diseases that can lead to severe illness or death. Furthermore, the debate often centers on the ethical responsibility of individuals to consider the health of their communities, particularly vulnerable populations. \\ \hline

\textbf{Key Differences}:  
Counter-Argument A emphasizes general vaccine safety, efficacy, and historical success (measles, polio, COVID-19), while Counter-Argument B focuses on ethics, mandates, and community health. 
In terms of addressing the label, A does not directly respond to the concern about ``mandatory'' vaccination and treats it as a broad safety issue, whereas B explicitly frames vaccination as a public health necessity, justifying mandates as a way to prevent outbreaks. 
The tone of A is neutral, scientific, and fact-based, centered on individual protection and evidence, while B balances acknowledgment of personal choice with collective responsibility, aligning more closely with the ``mandatory'' label. 
Regarding evidence, A cites the CDC, WHO, and past diseases as proof of vaccine safety, while B also references the CDC but ties its evidence specifically to mandates, herd immunity, and lives saved. 
Finally, the persuasive strategies differ: A appeals primarily to trust in science and institutions, whereas B combines scientific outcomes with ethical responsibility, making it more comprehensive for the specific concern. 
\\ \hline
\end{tabularx}
\caption{\centering Comparison Between Counter-Argument A and Counter Argument B}
\label{Table: Comparision between A and B}
\end{table*}

\begin{table*}[h!]
\centering
\renewcommand{\arraystretch}{1.2}
\setlength{\tabcolsep}{5pt}
\begin{tabularx}{\textwidth}{|l|X|}
\hline
\textbf{Tweet} & \texttt{@johnrobertsFox @AstraZeneca @KyraPhillips I'll take my chances with natural immunity by getting the virus and being treated with Hydroxichloroquine, zpak and zinc.} \\ \hline

\textbf{Human A} & Those treatments have no known benefit and there is no way to predict how serious your reaction to being infected could be, including death. \\ \hline

\textbf{Human B} & I would be careful taking a natural approach and then taking non-natural substances which might not be proven or recommended by health experts. \\ \hline

\textbf{GPT-4o-mini} & While the desire for natural immunity is understandable, relying on infection and unproven treatments like hydroxychloroquine poses significant risks. Natural infections can lead to severe illness, long-term health complications, and even death. Vaccines, on the other hand, provide a safer way to achieve immunity without the dangers of the virus itself. Additionally, the effectiveness of hydroxychloroquine has been widely discredited for COVID-19 treatment. Vaccination not only protects the individual but also helps build community immunity, reducing the overall spread of the virus. Prioritizing vaccines is a more reliable and safe approach to public health. \\ \hline

\textbf{Tiny-LLaMA} & While it is understandable to have concerns about potential side effects or risks from taking these supplements, it is important to consider that natural immunity may not provide as strong of a defense against COVID-19 as it does through vaccination. Additionally, widespread use of hydroxychloroquine and other treatments could help in controlling the spread of the virus and save lives during a global pandemic. It is also crucial to listen to exper \\ \hline

\textbf{Phi-3} & While it is important to consider all options when dealing with a health crisis like COVID-19, relying solely on one treatment method without proper guidance from medical professionals can be dangerous. Natural remedies may not always provide adequate protection against severe symptoms or long term effects of illnesses such as this pandemic has caused worldwide so far in comparison more research needs time. It’s best that people follow advice given out through credible sources rather than taking risks \\ \hline

\textbf{LLaMA-1B} & While it is important to consider all options when dealing with a potentially deadly disease like COVID-19, relying solely on natural immunity may not be effective in achieving full protection against severe illness or death from the virus. Additionally, there have been concerns about the efficacy of hydroxychloroquine in preventing transmission and reducing symptoms after exposure. It is crucial to consult with healthcare professionals before making decisions that could impact your health and safety. Consulting with medical experts can help you make an informed decision based on evidence rather than unproven methods. \\ \hline
\end{tabularx}
\caption{Comparison of human and model counter-arguments for a tweet promoting natural immunity and unproven COVID-19 treatments.}
\label{Table: human_annotations}
\end{table*}